\documentclass[10pt,twocolumn,letterpaper]{article}

\usepackage[pagenumbers]{iccv}      % To produce the REVIEW version

\definecolor{iccvblue}{rgb}{0.21,0.49,0.74}
\usepackage[pagebackref,breaklinks,colorlinks,allcolors=iccvblue]{hyperref}

\usepackage{color}
\usepackage{colortbl}
\usepackage{xcolor}
\usepackage{multirow}
\usepackage{makecell}
\usepackage{utfsym}
\usepackage{pifont}
\definecolor{deepgreen}{rgb}{0.0, 0.6, 0.0}
\newcommand{\cmark}{\scalebox{0.9}{\textcolor{deepgreen}{\ding{52}}}}
\newcommand{\xmark}{\textcolor{red}{{\ding{55}}}}
\usepackage{arydshln}

\def\paperID{2414} % *** Enter the Paper ID here
\def\confName{ICCV}
\def\confYear{2025}

\title{NuGrounding: A Multi-View 3D Visual Grounding Framework \\ in Autonomous Driving}

\author{
Fuhao Li$^{1,^*}$ \quad
Huan Jin$^{2,^*}$ \quad
Bin Gao$^{2}$ \quad
Liaoyuan Fan$^{3}$ \quad
Lihui Jiang$^{2}$ \quad
Long Zeng$^{1,^\dagger}$
\vspace{0.2em} \\
$^1$Tsinghua University \quad
$^2$Huawei Noah's Ark Lab \quad
$^3$The University of Hong Kong \\
{\tt\small lfh23@mails.tsinghua.edu.cn,
jinhuan3@huawei.com}
}

\begin{document}
\maketitle

\renewcommand\thefootnote{} %
\footnotetext{$^*$Equal contribution. $^\dagger$Corresponding author.}
\renewcommand\thefootnote{\arabic{footnote}} %

\begin{abstract}
Multi-view 3D visual grounding is critical for autonomous driving vehicles to interpret natural languages and localize target objects in complex environments. However, existing datasets and methods suffer from coarse-grained language instructions, and inadequate integration of 3D geometric reasoning with linguistic comprehension.
To this end, we introduce \textbf{NuGrounding}, the first large-scale benchmark for multi-view 3D visual grounding in autonomous driving. We present a Hierarchy of Grounding (HoG) method to construct NuGrounding to generate hierarchical multi-level instructions, ensuring comprehensive coverage of human instruction patterns.
To tackle this challenging dataset, we propose a novel paradigm that seamlessly combines instruction comprehension abilities of multi-modal LLMs (MLLMs) with precise localization abilities of specialist detection models. Our approach introduces two decoupled task tokens and a context query to aggregate 3D geometric information and semantic instructions, followed by a fusion decoder to refine spatial-semantic feature fusion for precise localization. Extensive experiments demonstrate that our method significantly outperforms the baselines adapted from representative 3D scene understanding methods by a significant margin and achieves 0.59 in precision and 0.64 in recall, with improvements of 50.8\% and 54.7\%.
\end{abstract}

\vspace{-1.0em}
\section{Introduction}
Multi-view 3D visual grounding plays a critical role in enabling autonomous driving vehicles \cite{uniad, fololane, ucc} to comprehend driving environments through natural language instructions.
This process involves the integration of multi-view images and textual instructions analysis into a unified 3D object localization framework, bridging the gap between human intent and machine perception. By facilitating human-centric scene understanding, it paves the way for safer and more intuitive human-vehicle interactions.

\begin{figure}[ht]
    \centering
    \includegraphics[width=0.48\textwidth]{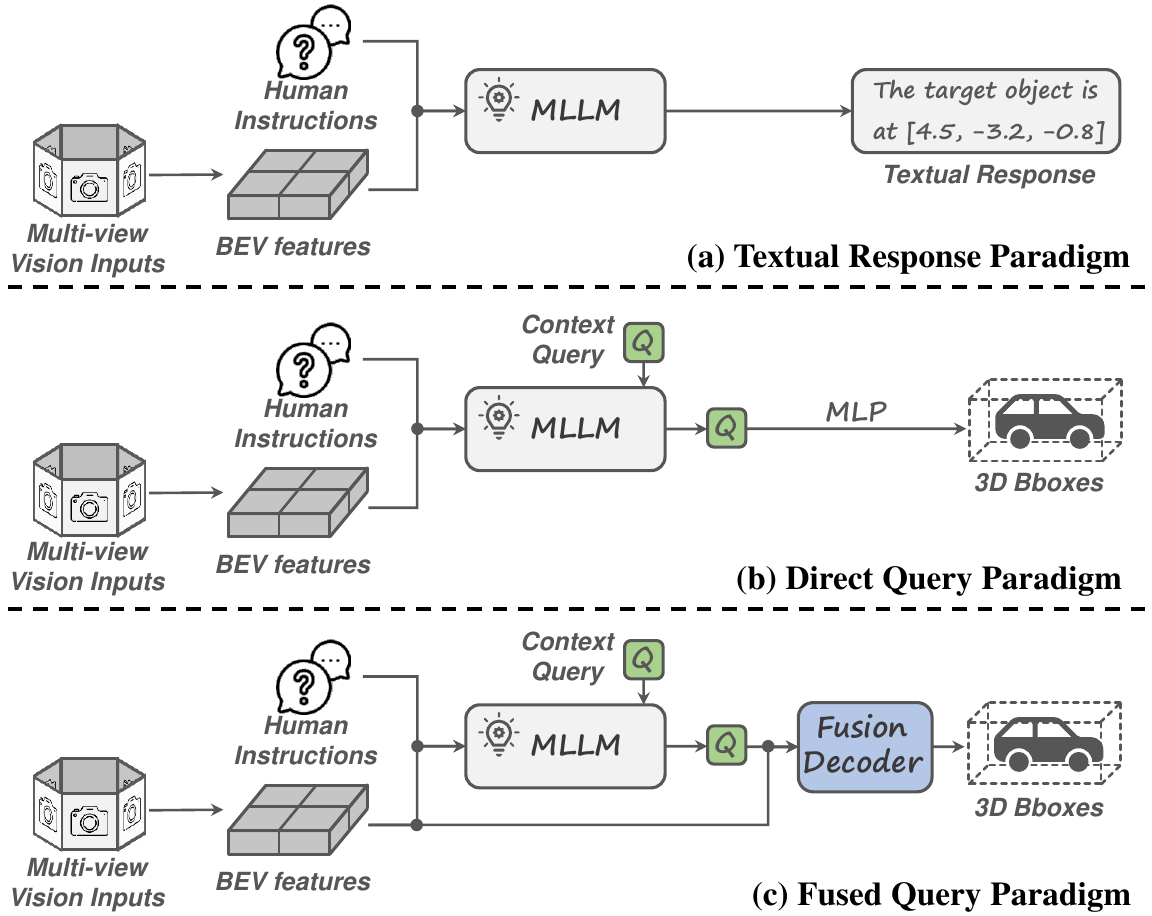}
    \caption{
    % Comparison of various MLLM-based multi-view 3D grounding frameworks. \textbf{(a)} Localize target objects by generating textual response. \textbf{(b)} Directly decode the MLLM hidden embeddings for 3D bounding box regression tasks via a learnable query. \textbf{(c) Ours:} Fuse the semantic context query with the 3D spatial details to generate the fused bounding boxes.
    \textbf{Comparison of various MLLM-based multi-view 3D grounding frameworks.} \textbf{(a)} Localize target objects by generating textual response. \textbf{(b)} Directly decode the MLLM hidden embeddings for predicting 3D bounding box simply. \textbf{(c) Ours:} Fuse the semantic context query with the 3D spatial information to for precise 3D bounding box regressions.
    }
    \vspace{-1.6em}
    \label{fig:framework}
\end{figure}

Although language-based autonomous driving systems have achieved remarkable progress, existing datasets are not suitable for multi-view 3D visual grounding due to oversimplified prompts, limited scale, and coarse-grained tasks. Prior visual grounding datasets \cite{refer-kitti, refer-kitti-v2, nuinstruct} focus solely on 2D pixel-level object localization without 3D geometric representations, while others \cite{mono3dvg, talk2car, llmi3d} concentrate on single-view image while neglecting holistic multi-view scene understanding. Furthermore, these datasets exhibit insufficient instruction diversity and quantity to cover generalized scenarios. Recent proposals \cite{nuscenesmqa,nuscenesqa,tod3cap,drivelm,omnidrive} primarily address scene-level tasks (e.g. visual question answering) or single-object description tasks (e.g. dense caption), failing for instance-level multi-object localization.

To address these gaps, we introduce \textbf{NuGrounding} dataset, the first large-scale benchmark designed for multi-view 3D visual grounding in autonomous driving. Different from prior works, NuGrounding supports multi-object references, instance-level localization, and balanced prompt complexity and quantity. To construct the dataset, we collect object attributes in NuScenes \cite{nuscenes} through automatic labeling and little manual verification. We then propose the Hierarchy of Grounding (\textbf{HoG}) method to generate hierarchical multi-level textual prompts.

The multi-view 3D visual grounding task requires both complex human instruction and fine-grained scene understanding. Numerous attempts had been made prior to this work.
% Inspired by the multi-modal information processing capability of MLLMs, we aim to leverage MLLMs to address this multi-view 3D visual grounding challenges. 
% However, existing MLLMs for remain suboptimal for this task. 
As shown in Fig. \ref{fig:framework}(a), prior studies \cite{3d-tokenizer, omnidrive, nuinstruct} typically encode multi-view images into BEV features and integrate 3D spatial reasoning capabilities into MLLMs. However, these approaches primarily concentrate on response generation, which limits their effectiveness in fine-grained object localization. As shown in Fig. \ref{fig:framework}(b), Recent methods \cite{llmi3d} attempt to decode LLM hidden embeddings into 3D bounding box regression tasks via 3D queries. However, these 3D queries reside in semantic embedding space and lack fine-grained 3D geometric details, thereby hindering precise 3D localization.

To this end, we propose a multi-view 3D visual grounding framework, a novel paradigm that seamlessly combines the instruction comprehension capabilities of MLLMs with the precise object localization abilities of specialist detection models, as shown in Fig. \ref{fig:framework}(c). Specifically, firstly we employ a BEV-based detector to extract dense BEV features and generate instance-level object query with 3D geometric priors. Secondly, We decouple the alone task token \cite{lisa} into a textual prompt token and a downstream embedding token, which helps our predefined context query to aggregate 3D geometric information and semantic instructions. Finally, we introduce a fusion decoder to integrate the semantic information with the 3D spatial details to predict bounding boxes. The framework achieves both complex human instruction understanding and multi-view scene perception for accurate object localization.

Overall, our contributions can be summarized as follows:
\begin{itemize}
    \item We introduce \textbf{NuGrounding} dataset, the first large-scale dataset for multi-view 3D visual grounding in autonomous driving. To ensure diversity, scalability, and generalization, we propose the Hierarchy of Grounding (HoG) method to construct NuGrounding.
    \item We propose the multi-view 3D grounding framework, which is a novel paradigm that seamlessly combines the instruction comprehension capabilities of MLLMs with the precise object localization abilities of specialist detection models.
    \item We adapt state-of-the-art methods and evaluate them on our NuGrounding dataset, establishing a comprehensive benchmark. The experiments demonstrate our method significantly outperforms adapted baselines, achieving improvements of 50.8\% in precision and 54.7\% in recall.
\end{itemize}
\belowrulesep=0pt
\aboverulesep=0pt

\section{Related Work}

\begin{table*}[t]
    \centering
    \resizebox{0.8\textwidth}{!}{
    
    \begin{tabular}{c||cccc|ccc}
    \toprule[1.25pt]
    \multirow{2}{*}{Dataset} & \multirow{2}{*}{3D} & \multirow{2}{*}{Multi-view} & \multirow{2}{*}{\shortstack{Object \\ Detection}} & \multirow{2}{*}{Multi-object} & \multirow{2}{*}{Prompts} & \multirow{2}{*}{Frames} & \multirow{2}{*}{\shortstack{Instances \\ per Prompt}} \\
     &  &  &  &  &  &  \\ \midrule
    Refer-KITTI\cite{refer-kitti} & \xmark & \xmark & \cmark & \cmark & 0.8K & 6650 & 10.7 \\
    Mono3DRefer \cite{mono3dvg} & \cmark & \xmark & \cmark & \xmark & 41K & 2025 & 1 \\
    Talk2Car \cite{talk2car} & \cmark & \xmark & \cmark & \xmark & 12K & 9217 & 1 \\
    NuInstruct \cite{nuinstruct} & \cmark & \cmark & \xmark & \cmark & 91K & 11850 & - \\
    Nuprompt \cite{nuprompt} & \cmark & \cmark & \xmark & \cmark & 35K & 34149 & 5.3 \\
    NuscenesQA \cite{nuscenesqa} & \cmark & \cmark & \xmark & \xmark & 0.5M & 34149 & - \\ 
    TOD$^{3}$Cap \cite{tod3cap} & \cmark & \cmark & \xmark & \xmark & 2.3M & 34149 & 1 \\ 
    OmniDrive \cite{omnidrive} & \cmark & \cmark & \xmark & \cmark & 0.2M & 34149 & - \\ \midrule
    \textbf{NuGrounding} & \cmark & \cmark & \cmark & \cmark & 2.2M & 34149 & 3.7 \\
    \bottomrule[1.25pt]
    \end{tabular}
    
    }
    \caption{\textbf{Comparison of NuGrounding (ours) with existing language-based driving datasets.} Multi-object denotes multiple objects for detection, not caption. - means no object detection ability. NuGrounding provides comprehensive information and complex, substantial prompt for comprehensive autonomous driving understanding.}
    \label{table:dataset}
\end{table*}

\subsection{Multi-View 3D Detection}
Multi-view 3D detection has gained significant interest for autonomous driving. The LSS series \cite{lss, bevdet, bevdepth, bevdet4d} project image features into BEV space via pixel-level depth prediction, and directly decode BEV feature to objects. In contrast, others utilize the sparse object query for object regression. DETR-based \cite{petr, streampetr} methods employ object queries to decode bounding box directly. However, these methods lack language guidance for cross-modal scene understanding. Our approach seamlessly integrates language guidance into query-based detection models.

\subsection{Visual Understanding Datasets for Driving}
Numerous existing datasets have significantly advanced visual grounding in autonomous driving \cite{occclip, foundationmodels}. A comprehensive comparison between benchmark datasets and ours is summarized in Tab. \ref{table:dataset}. Building upon KITTI \cite{kitti}, Refer-KITTI \cite{refer-kitti} first established a referring multi-object tracking dataset but focuses solely on 2D bounding boxes of objects. Subsequently, Mono3DRefer \cite{mono3dvg} and Talk2Car \cite{talk2car} correspond objects with 3D geometric attributes for grounding but concentrate only on monocular information. NuPrompt \cite{nuprompt} proposed the first multi-view object-centric language dataset, but exhibit insufficient prompt diversity and quantity. Recently, NuScenesQA \cite{nuscenesqa} and NuScenesMQA \cite{nuscenesmqa} expand the data quantity but focus on scene-level VQA rather than instance-level localization. Similarly, TOD$^{3}$Cap \cite{tod3cap} is designed for dense captioning and restricts each textual prompt to a single object, limiting its applicability in multi-referred-objects scenarios. In contrast, we propose the first large-scale 3D visual grounding dataset NuGrounding for multi-view autonomous driving.

\subsection{Visual Understanding Models for Driving}
Traditional multi-view 3D grounding method \cite{nuprompt} can only deal with simple texts but fail to understand complex human instructions. Recently, with the development of multi-modal LLMs \cite{llava, minigpt, internvl}, many autonomous driving models \cite{drivelm,drivevlm,drivegpt4} leverage MLLM to comprehend the driving scenes. However, these methods only generate scene-level textual outputs, which is not fit for instance-level localization. Others \cite{3d-tokenizer,nuinstruct,elm,omnidrive} utilize MLLMs to get the spatial location of text-guided objects of interest. However, they output the results by generating textual responses rather than regressing bounding boxes. LLMI3D \cite{llmi3d} attempts to bridge LLM text generation with bounding boxes regression by a 3D query and task token mechanism. However, this 3D query is in semantic embedding space without geometric details, preventing accurate 3D localization. In contrast, our framework combines both instruction understanding and precise localization.

\begin{figure}[ht]
    \centering
    \includegraphics[width=0.48\textwidth]{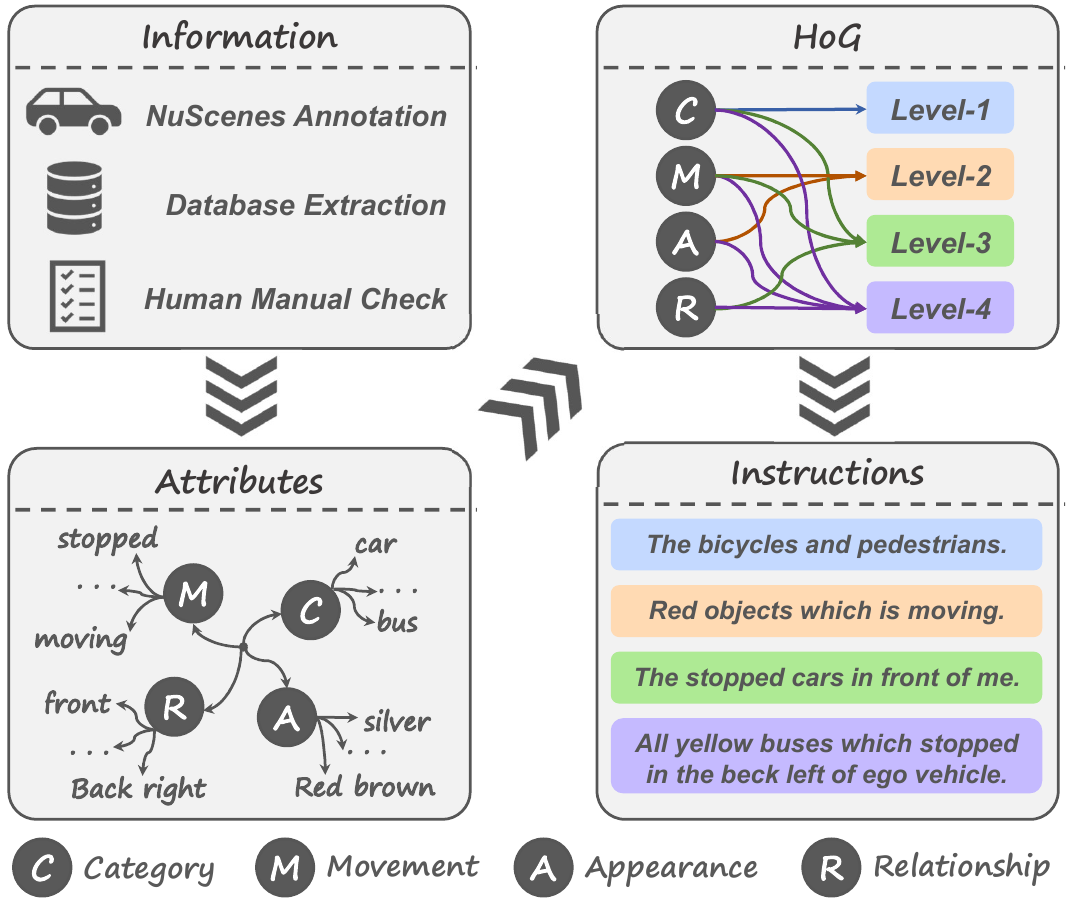}
    \caption{
    \textbf{Data construction flow of NuGrounding.} First, we annotate diverse common attributes for each object. The category, movement and relationship attributes are generated by rule-based calculations using the official annotated information. The appearance is extracted from other datasets, and we manually check its correction. Then the proposed Hierarchy of Grounding (HoG) method is used to generate textual prompt across four levels.
    }
    \label{fig:pipeline}
\end{figure}

\section{NuGrounding Dataset}

\begin{figure*}[t]
    \centering
    \includegraphics[width=1.0\linewidth]{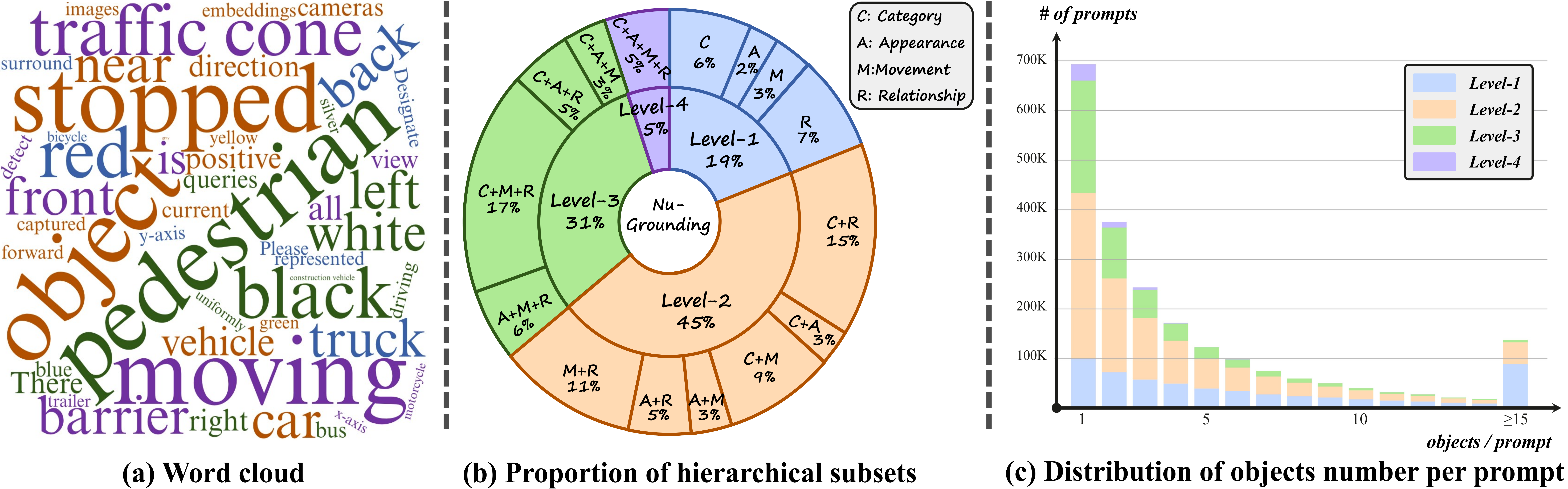}
    % \vspace{-2.0em}
    \caption{Statistics of NuGrounding dataset. \textbf{(a) Word cloud.} It represents the top 50 words used in NuGrounding. \textbf{(b) Proportion of hierarchical subsets.}  NuGrounding can be divided into four levels based on the number of selected attribute combinations. The size of the arc represents the proportions of each subset, while the same color indicates subsets of the same level. \textbf{(c) Distribution of objects number per prompt.} The horizontal axis represents the amount of objects corresponding to each prompt, while the vertical axis represents the amount of this kind of prompts.
    }
    \label{fig:statistics}
\end{figure*}

Existing visual understanding datasets for driving are not suitable for multi-view 3D visual grounding due to oversimplified prompts, limited scale, and coarse-grained tasks.
To address these gaps, we proposed the first multi-view 3D visual grounding dataset, \textbf{NuGrounding}, which is constructed from NuScenes \cite{nuscenes}. Additionally, we propose the Hierarchy of Grounding (HoG) method to generate hierarchical multi-level textual prompts. Specifically, as illustrated in Fig. \ref{fig:pipeline}, we first annotate diverse common attributes for each object (Sec. \ref{section:attr_collection}). Then fill these attributes into HoG method to obtain textual instructions (Sec. \ref{section:prompt_construction}).

\subsection{Attribute Collection}
\label{section:attr_collection}
When describing specific objects by natural language, humans typically focus on their intrinsic attributes: category, appearance, movement, and spatial relationship relative to the ego vehicle \cite{egothink}. Accordingly, we perform instance-level annotations for these four attribute types.

\textbf{Category:} We adopt ten common object categories in NuScenes: \textit{car}, \textit{truck}, \textit{bus}, \textit{trailer}, \textit{construction vehicle}, \textit{pedestrian}, \textit{motorcycle}, \textit{bicycle}, \textit{barrier}, and \textit{traffic cone}.

\textbf{Appearance:} Recognizing object appearance, particularly vehicle colors in driving scenarios, is critical. NuPrompt \cite{nuprompt} manually annotates color information in video sequences, TOD$^{3}$Cap \cite{tod3cap} employs pre-trained caption models for automated color extraction. We merge color annotations from these two datasets and perform meticulous manual verification for inconsistent annotations.

\textbf{Movement:} Accurate movement estimation is therefore essential for scene understanding and grounding. We calculate inter-frame displacement to estimate object velocities, categorizing motion states as \textit{moving} or \textit{stopped} using $0.3 m/s$ threshold.

\textbf{Relationship:} Following NuScenesQA \cite{nuscenesqa}, we define six relationships corresponding to six camera views: \textit{front}, \textit{front left}, \textit{front right}, \textit{back}, \textit{back left}, and \textit{back right}. Each relationship covers a 60° field of view (FOV) in the bird’s-eye-view (BEV) plane to ensure unique assignment.

\subsection{Hierarchy of Grounding Construction}
\label{section:prompt_construction}
After collecting instance-level attributes, we combine them using the proposed Hierarchy of Grounding \textbf{(HoG)} method to generate scene-level textual prompts. The HoG method not only accommodates all kinds of human instructions, but also prevents inductive bias.

First, humans tend to describe a group of objects using their shared attributes (e.g., ``\textit{pedestrians around me}"), but locate the specific object through unique attribute combinations (e.g., ``\textit{the moving orange bus in front left of me}"). The number of combined attributes correlates with the referential specificity and the prompt complexity. This motivates our hierarchical prompt generation strategy at multiple difficulty levels by controlling the number of stacked attributes, in order to cover human description patterns more generally.

\label{sec:HoG}
Second, expressions coupling all four attribute types without hierarchy may lead to inductive bias \cite{eda}. For instance, in a scene with only one car, prompts like ``\textit{the moving red car in my front left}" and ``\textit{the car}" refer to the same object. Training on a large number of such samples would guide the model to take shortcuts to only focus on the category attribute while neglecting other attributes, resulting in deviant learning. This emphasizes the importance of attribute decoupling and multi-level hierarchical construction. See supplementary Sec. \ref{sec:HoG_motivation} for detail visualization.

Concretely, we generate $C_{4}^{1} + C_{4}^{2} + C_{4}^{3} + C_{4}^{4} = 15$ templates by selecting varying attribute combinations. Templates picking $N$ attribute types are termed $N$-level prompts. Then we traverse all the picked attributes of the objects in the current scene and fill them into the templates to obtain semantic expressions, as shown in Fig. \ref{fig:pipeline}.

\begin{figure*}[t]
    \centering
    \includegraphics[width=1.0\linewidth]{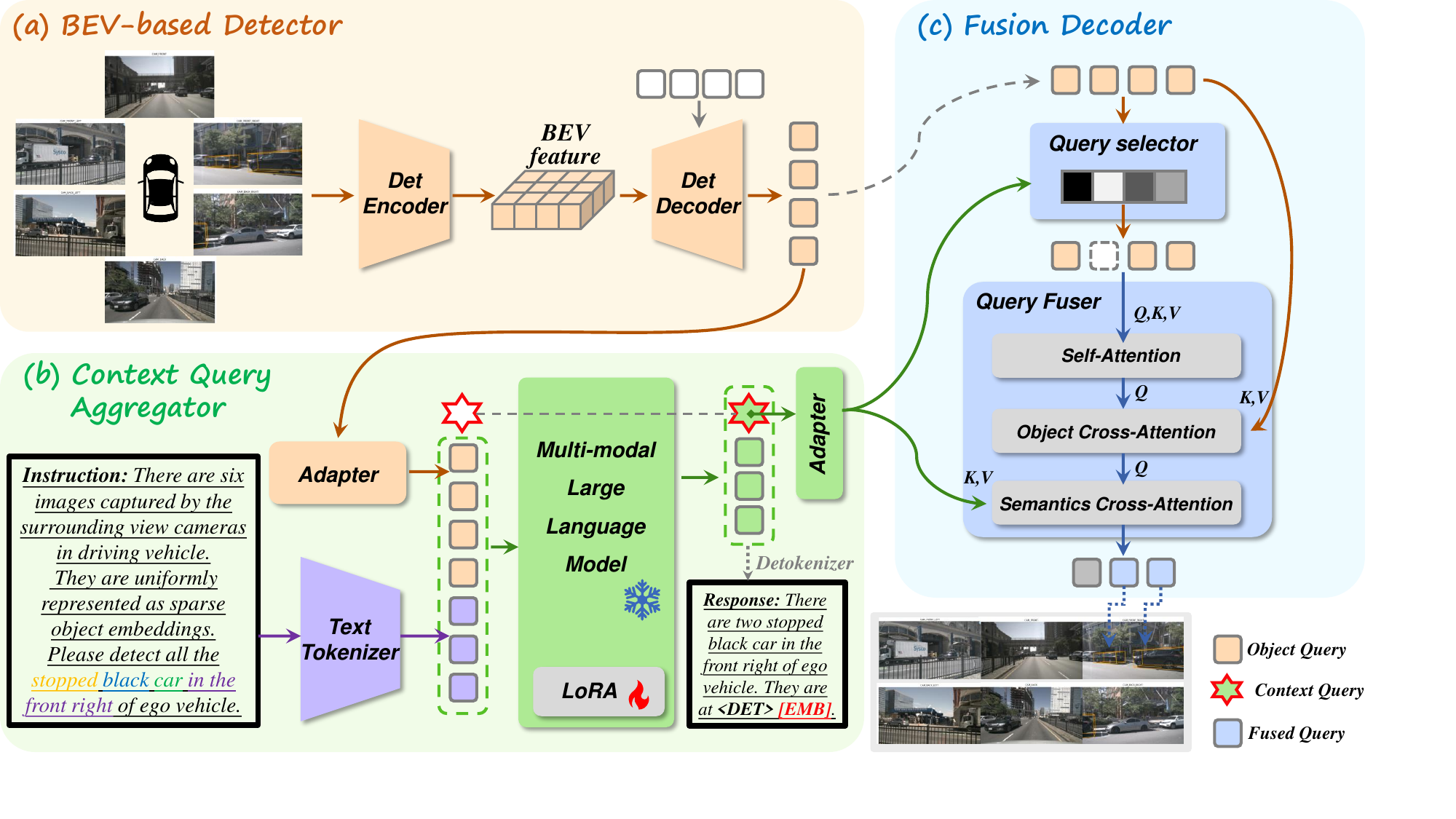}
    % \vspace{-2.0em}
    \caption{
    \textbf{Overall architecture of the proposed NuGrounding.} It consists of three parts: a bev-based detector that provides visual embedding, a context query aggregator designed to accommodate both visual embedding and language instruction, and a fusion decoder that fuses the semantic context query and the 3d object query.
    }
    \label{fig:method}
\end{figure*}

\subsection{Data Statistics}

NuGrounding samples 34,149 key frames from 850 videos within NuScenes, generating 2.2M textual prompts (63.7 prompts per frame). The dataset is divided into 1.8M prompts from 28,130 frames for training and 0.4M prompts for testing. Crucially, NuGrounding hierarchically generates prompts with varied attribute combinations to balance difficulty levels. As shown in Fig. \ref{fig:statistics}~(b), each hierarchical subset maintains approximately equal proportion, preventing models from taking textual shortcuts and improving generalization. Furthermore, NuGrounding supports multiple referred objects per prompt, with each prompt referencing an average of 3.7 objects shown in Fig. \ref{fig:statistics}~(c).
\section{Methodology}

The 3D visual grounding task in driving scenarios requires comprehensive capabilities of \ding{182} multi-view scene perception, \ding{183} complex human instruction understanding, \ding{184} precise 3D object localization. However, existing 3D detection models lack the capabilities to comprehend human instructions, while MLLMs suffer from low precision for object localization. To this end, we propose a novel paradigm that combines the instruction understanding of MLLMs with the precise object localizations of specialist detection models. Our framework achieves both complex human instruction understanding and accurate object localization.

\subsection{Overall Architecture}
The architecture of our method is illustrated in Fig.\ref{fig:method}. Firstly, in \textbf{BEV-based Detector}, a specialist detection encoder is employed to extract dense BEV features from multi-view images, followed by a query-based detection decoder to utilize the extracted feature to generate the sparse instance-level object queries (Sec. \ref{sec:detector}). Secondly, in \textbf{Context Query Aggregator}, the object queries serve as sparse scene representations to be fed into the MLLM along with textual instructions. Additionally, we introduce two separate task token together with a learnable context query. While generating textual responses, the MLLM follows the task tokens to aggregates the 3D scene information and textual instructions into this context query (Sec. \ref{sec:aggregator}). Finally, in \textbf{Fusion Decoder}, the object queries are filtered based on their relevance to the context query to eliminate semantically irrelevant instance-level noise. The selected queries then enhance their spatial information through interactions with all object queries, and integrate semantic information through interactions with the context query, ultimately generating the fused queries. Finally, these fused queries are decoded by the specialist object decoder (Sec. \ref{sec:decoder}).

\subsection{BEV-based Detector}
\label{sec:detector}
Following multi-view 3D detection methods \cite{streampetr, bevformer}, the BEV-based detector extracts multi-view image information by constructing the BEV feature and transforms it into instance-level object queries.

\textbf{Detection Encoder.} Firstly, multi-view images are fed into an image backbone network \cite{vovnet, ViT} to extract the image features. Subsequently, according to the projection matrix of cameras, the image features are transformed to the grid-shaped BEV plane to construct the BEV features.

\textbf{Detection Decoder.} We initialize a set of learnable 3D anchors as object queries $Q_{obj} \in \mathbb{R}^{N_{obj} \times C_{B}}$, where $N_{obj}$ is the predefined number of object queries. These sparse $Q_{obj}$ is fed into a transformer architecture to aggregate useful dense BEV information.

\subsection{Context Query Aggregator}
\label{sec:aggregator}
After obtain the object query, we consider it as the sparse scene information and feed it to MLLM together with human instructions. Subsequently, a context query is seamlessly integrated into the MLLM inference process to aggregate 3D scene information and textual instruction information. finally The MLLM outputs both the text response and the aggregated context query.

\textbf{Multi-modal Input.} Current MLLMs \cite{llava, minigpt, visionllmv2, internvl} and LLM-based autonomous driving models \cite{elm, drivevlm, drivelm} utilize 2D features extracted from images as visual input without incorporating 3D geometric priors. In contrast, we use sparse object queries $Q_{obj}$ as 3D scene representations for the MLLM. To bridge the inherent gap between 3D scene information and widely pretrained 2D MLLMs, following \cite{3d-tokenizer, omnidrive}, a two-layer linear adapter $\gamma_{1}$ is employed to align the 3D features with the LLM. The aligned 3D features are then combined with the tokenized text features $x_{txt}$ as the whole multi-modal input $x_{m}$. It can be formulated as:
\begin{equation}
    x_{m} = Concat([\gamma_{1}(Q_{obj}), x_{txt}])
\end{equation}

In addition, unlike which utilize dense BEV features \cite{tod3cap}, sparse object queries are intuitively more compatible with the LLM input pattern. Each token in the object query set represents an independent representation of an object of interest within the 3D scene, with inherent spatial correlations among tokens.

\textbf{Context Query.} This step aims to aggregate the multi-modal input $x_{m}$ and distill its effective information. Most LLM-based models \cite{segpoint,3d-llava,llava-3d} follow LISA \cite{lisa} to introduce a task token mechanism for multi-modal input aggregation. However, the alone task token of all these methods is responsible for both downstream task decoding and textual answer generation. This trade-off will potentially hinders multi-modal information aggregation. Therefore, we decouple the alone task token into a task symbol token which signals that the next token will be used for aggregation, and a downstream placeholder token whose word embedding will be replaced by the predefined context query to aggregate multi-modal information.

Specifically, we first initialize a learnable parameter as the context query $Q_{cont} \in \mathbb{R}^{1 \times C_{L}}$, where $C_{L}$ is the dimension of LLM last hidden layer. Next, we add two new tokens, e.g., \texttt{[DET]} and \texttt{[EMB]}, to the original LLM vocabulary. During the LLM response generation process, the \texttt{[EMB]} token always follows behind \texttt{[DET]}. The \texttt{[DET]} token serves as a task symbol to signal that the subsequent token will be used for multi-modal information aggregation. The subsequently generated \texttt{[EMB]} token is masked to \textbf{be excluded from calculating cross-entropy loss} for text generation, and its corresponding word embedding is replaced by the predefined context query for the following auto-regressive procedure. Consequently, the context query operates within this auto-regressive mechanism to effectively aggregate both 3D scene and textual semantic information. The process can be formulated as:
\begin{gather}
x_{a}^{n+1} = \mathcal{F}_{LLM}([x_{m}; x_{a}^{1}, x_{a}^{2}, \cdot \cdot \cdot,x_{a}^{n}]), \\
{\rm \texttt{[DET]}} = De(x_{a}^{n+1}), \\
\tilde{Q}_{cont} = \mathcal{F}_{LLM}([x_{m};x_{a}^{1}, \cdot \cdot \cdot,x_{a}^{n}, Q_{cont}])
\end{gather}
Where $\mathcal{F}_{LLM}$ is the auto-regressive mechanism in LLM, $\{x_{a}^i|i=1,2,\cdot \cdot \cdot,n+1\}$ is the response token feature, and $n$ is the turn when meeting \texttt{[DET]} token translated by detokenizer $De$.

\textbf{Textual Thinking.} While the context query aggregates multi-modal information, the MLLM simultaneously generates the text response. Most LLM-based approaches \cite{lisa, llmi3d, 3d-llava} design a very simple response format (e.g., \textit{It is [DET]}). However they neglect the potential of textual responses to guide the context query performing thinking and understanding for multi-modal inputs. In fact, in the LLM auto-regressive process, the context query not only interact with the multi-modal input but also with all response tokens generated before \texttt{[DET]}. We find that the scene and target object descriptions in textual response significantly influences the aggregation effectiveness. We figure that the detailed textual response enables MLLM to take more time and more response tokens for reasonable thinking during inference \cite{emerg-thinking}. This thinking process increases the model complexity and representation ability and makes the context query understand the whole multi-modal information and aggregate more useful information. Therefore, we provide detailed target object descriptions in the response, such as \textit{There are two stopped black car in the front right of ego vehicle. They are at [DET] [EMB]}.

\begin{table*}[ht]
    \centering
    \renewcommand\arraystretch{1.2}
    % \caption{\textbf{Multi-view 3D visual grounding results among NuGrounding (ours) and previous related works.} We adapt other methods on our proposed NuGrounding dataset with the same configuration for fair comparison. P and R mean Precision and Recall, respectively.}
    \resizebox{\textwidth}{!}{
    \begin{tabular}{c|c|cccc|cccc|cccc|cccc|cccc}
    \toprule[1.25pt]
    \multirow{2}{*}{Method} & \multirow{2}{*}{Backbone} & \multicolumn{4}{c|}{Level-1} & \multicolumn{4}{c|}{Level-2} & \multicolumn{4}{c|}{Level-3} & \multicolumn{4}{c|}{Level-4} & \multicolumn{4}{c}{Average} \\ \cmidrule{3-22}
    & & P & R & mAP & NDS & P & R & mAP & NDS & P & R & mAP & NDS & P & R & mAP & NDS & P & R & mAP & NDS \\
    \midrule
    ELM (ECCV24)             & ViT-L                     & 0.19          & 0.14          & -             & -             & 0.05          & 0.02          & -             & -             & 0.08          & 0.06          & -             & -             & 0.07          & 0.09          & -             & -             & 0.07          & 0.08          & -             & -              \\
    NuPrompt (AAAI25)        & V2-99                     & 0.33          & 0.23          & 0.09             & 0.20             & 0.20          & 0.16          & 0.05             & 0.16             & 0.21          & 0.27          & 0.08             & 0.17             & 0.23          & 0.29          & 0.09             & 0.19             & 0.22          & 0.24          & 0.08             & 0.18              \\
    NuPrompt (AAAI25)        & ViT-L                     & 0.37          & 0.27          & 0.13             & 0.24             & 0.23          & 0.25          & 0.08             & 0.18             & 0.29          & 0.32          & 0.12             & 0.20             & 0.26          & 0.34          & 0.11             & 0.21             & 0.29          & 0.29          & 0.11             & 0.22              \\
    OmniDrive (CVPR25)       & ViT-L                     & 0.22          & 0.16          & -             & -             & 0.24          & 0.30          & -             & -             & 0.11          & 0.08         & -             & -             & 0.12         & 0.15          & -             & -             & 0.17          & 0.17          & -             & -              \\
    \midrule
    \multirow{2}{*}{\textbf{NuGrounding}}    & V2-99                     & 0.46          & 0.57          & 0.31          & 0.47          & 0.49          & 0.52          & 0.30          & 0.40          & 0.51          & 0.53          & 0.32          & 0.42          & 0.47          & 0.54          & 0.31          & 0.41          & 0.49          & 0.54          & 0.31          & 0.42           \\
    & ViT-B                     & \textbf{0.64} & \textbf{0.72} & \textbf{0.47} & \textbf{0.57} & \textbf{0.57} & \textbf{0.60} & \textbf{0.37} & \textbf{0.44} & \textbf{0.58} & \textbf{0.62} & \textbf{0.40} & \textbf{0.47} & \textbf{0.56} & \textbf{0.61} & \textbf{0.38} & \textbf{0.45} & \textbf{0.59} & \textbf{0.64} & \textbf{0.40} & \textbf{0.48} \\
    \bottomrule[1.25pt]
    \end{tabular}     
    }
    \caption{\textbf{Multi-view 3D visual grounding results among NuGrounding (ours) and previous related works.} We adapt other methods on our proposed NuGrounding dataset with the same configuration for fair comparison. P and R mean Precision and Recall, respectively.}
    \label{tab:compareSOTA}
\end{table*}

\begin{table*}[t]
% \vspace{-2em}
\centering
%#################################################
%#################################################
\subfloat[
\textbf{Ablation on the components of the framework.}
\label{tab:components}
]{
\begin{minipage}{0.5\linewidth}{
\begin{center}
% \tablestyle{2.0pt}{1.0}
% \hspace{-1cm}
% \resizebox{\textwidth}{15mm}{
\scalebox{1.0}{
\setlength{\tabcolsep}{3pt}
\begin{tabular}{ccc|cc}
                \toprule[1.25pt]
                \begin{tabular}[c]{@{}c@{}}Query Fuser\end{tabular}  & \begin{tabular}[c]{@{}c@{}}Context Query\end{tabular} & \begin{tabular}[c]{@{}c@{}}Query Selector\end{tabular} & mAP & NDS \\
                \midrule
                \usym{2713} &  &  &  0.387 & 0.445 \\
                \usym{2713} & \usym{2713} &  &  0.443 & 0.497 \\
                \usym{2713} & \usym{2713} & \usym{2713} &  \textbf{0.458} & \textbf{0.505} \\
                \bottomrule[1.25pt]
                \end{tabular} %}
}
\end{center}
}
\end{minipage}
}
%\hspace{1mm}
%#################################################
%#################################################
\subfloat[
\textbf{Ablation study on the scene representation of visual input.}
\label{tab:scene_representation}
]{
\begin{minipage}{0.5\linewidth}{\begin{center}
% \tablestyle{2.0pt}{1.0}
% \hspace{-0.5cm}
\scalebox{1.0}{
\begin{tabular}{c|cc}
                \toprule[1.25pt]
                Scene Representation & mAP & NDS \\
                \midrule
                Blind & 0.307 & 0.389 \\
                BEV Feature & 0.345 & 0.418 \\
                Object Query & \textbf{0.458} & \textbf{0.505} \\
                \bottomrule[1.25pt]
                \end{tabular}
}
% \hspace{-10mm}
\end{center}
}
\end{minipage}
} \\
%#################################################
%#################################################
\subfloat[
\textbf{Ablation study on the detail description and information of textual thinking.} Cat. means category. Rel. means relationship. 
\label{tab:context_learning}
]{
\begin{minipage}{0.5\linewidth}{
\begin{center}
% \tablestyle{2.0pt}{1.0}
% \hspace{-1cm}
\scalebox{1.0}{
\begin{tabular}{ccc|cc}
                \toprule[1.25pt]
                \multicolumn{3}{c|}{Textual Thinking} & \multirow{2}{*}{mAP} & \multirow{2}{*}{NDS} \\ \cmidrule{1-3}
                Number & Cat. \& Rel. & Movement &  &  \\
                \midrule
                \usym{2717} & \usym{2717} & \usym{2717} & 0.387 & 0.445 \\
                \usym{2713} &  &  & 0.404 & 0.452 \\
                \usym{2713} & \usym{2713} &  & \textbf{0.458} & 0.505 \\
                \usym{2713} & \usym{2713} & \usym{2713} & 0.456 & \textbf{0.516} \\
                \bottomrule[1.25pt]
                \end{tabular}
}
\end{center}
}
\end{minipage}
}
%#################################################
%#################################################
\subfloat[
\textbf{Ablation study on the query number of query selector.}
\label{tab:QuerySelector}
]
{
\begin{minipage}{0.5\linewidth}{
\begin{center}
% \tablestyle{2.0pt}{1.0}
% \hspace{-1cm}
\scalebox{1.0}{
\begin{tabular}{c|cc}
                \toprule[1.25pt]
                Query Number & mAP & NDS \\
                \midrule
                32 & 0.309 & 0.293 \\
                64 & 0.439 & 0.493 \\
                256 & \textbf{0.458} & \textbf{0.505} \\
                900 (full) & 0.443 & 0.497 \\
                \bottomrule[1.25pt]
                \end{tabular}
}
\end{center}
}
\end{minipage}
}
\hspace{0.5mm}
%#################################################
%#################################################

\vspace{-5pt}
\caption{\textbf{A set of ablative studies on multi-view 3D visual grounding task.} Including the components of the framework, the scene representation of visual input, the details of textual thinking, and the query number of query selector. More details can be seen in Sec. \ref{sec:Ablation Study}.}
% ``BF.'': BEVFormer \cite{li2022bevformer}; ``BD.'': BEVDepth \cite{li2023bevdepth}.

% \vspace{-20pt}
\label{tab:ablation}
% \underfigtab
\end{table*}
% \hspace{-18pt}

\subsection{Fusion Decoder}
\label{sec:decoder}
During the auto-regressive process, the context query is mapped to the semantic embedding space as the LLM last layer’s output. As a result, the aggregated context query $\tilde{Q}_{cont}$ is rich in semantic and scene understanding but lacks 3D spatial information and fine-grained geometric details. Conversely, the object query $Q_{obj}$ extracted from the BEV-based detector has abundant 3D geometric priors but lacks semantic instruction guidance. Therefore, we design a fusion Decoder to integrate the semantic information from $\tilde{Q}_{cont}$ with the 3D geometric information from $Q_{obj}$ to generate fused query $Q_{fusion}$, which are subsequently decoded to predict 3D bounding boxes.

\textbf{Query Selector.} The purpose of the Query Selector is to filter out the most semantically relevant object queries $Q_{obj}^*$, thereby eliminating irrelevant object noise. Most selection methods \cite{groundingdino,lavida, glip} use multiple discrete text tokens to calculate token-level similarity, risking concentrating on semantically unrelated text tokens (e.g. \textit{moving} in the sentence \textit{the red bus that is not moving}) and selecting semantically ambiguous objects.
In contrast, we consolidate the entire semantic information into a single context query to calculate semantics-level similarity. This approach concentrates on globally semantic information during selection, thereby avoiding and ensuring deep semantic consistency.

Specifically, we first apply two MLP to both context query and object query to align them into a unified space. Then, we calculate the cosine similarity between them to obtain a similarity matrix $M_{sim}$ which measures semantic relevance. Finally, we select the top k object queries based on semantic similarity. The process can be formulated as:
\begin{gather}
\hat{Q}_{obj}=MLP_1(Q_{obj}),\enspace \hat{Q}_{cont}=MLP_2(\tilde{Q}_{cont}), \\
M_{sim}(\hat{Q}_{obj},\hat{Q}_{cont})=\frac{\hat{Q}_{obj}\cdot {\hat{Q}_{cont}^\texttt{T}}}{||\hat{Q}_{obj}||\cdot ||\hat{Q}_{cont}||} \\
Q_{obj}^*={Q}_{obj}[TopK(M_{sim},k)]
\end{gather}

\textbf{Query Fuser.} The Query Fusion Module employs a transformer architecture to perform cross-modal query fusion. Specifically, the selected object queries $Q_{obj}^*$ are sequentially fed into a self-attention layer, an object cross-attention layer, and a semantics cross-attention layer. In the object cross-attention layer, $Q_{obj}^*$ interacts with the original sparse scene representation $Q_{obj}$ to enhance their relative spatial position. In the semantics cross-attention layer, $Q_{obj}^*$ enhance their semantic information from $\tilde{Q}_{cont}$. Finally, the fused queries are used to predict 3D bounding boxes.

\belowrulesep=0pt
\aboverulesep=0pt

\section{Experiment}
\subsection{Experimental Setting}

\begin{figure*}[t]
    \centering
    \includegraphics[width=1.0\linewidth]{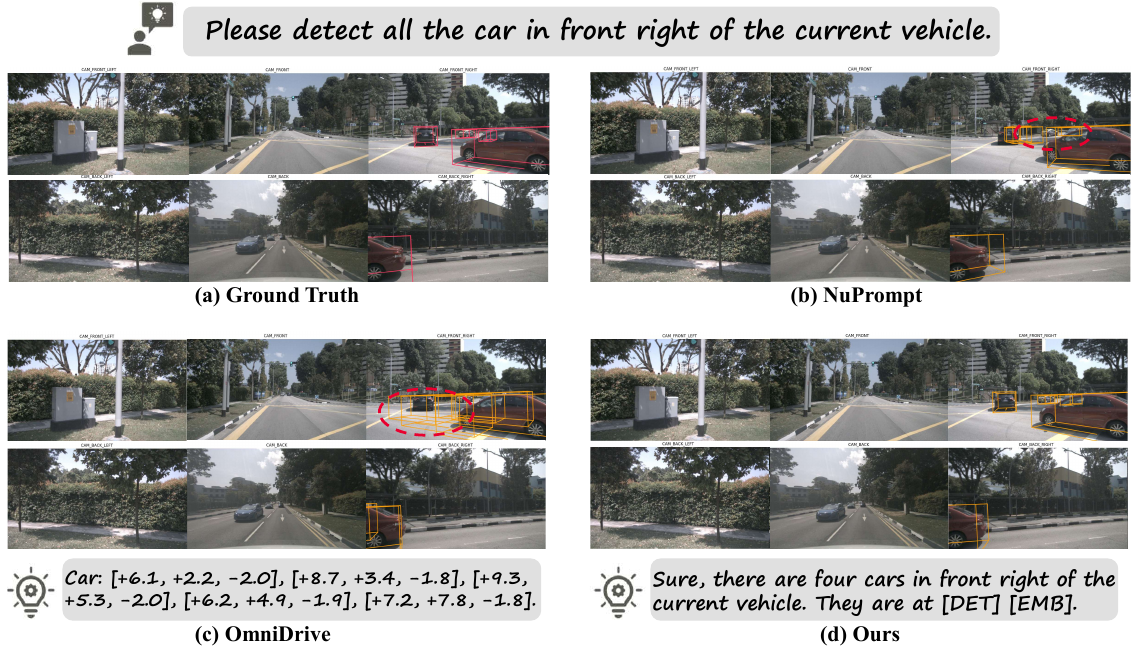}
    % \vspace{-2.0em}
    \caption{
    \textbf{Visual comparison among NuGrounding (ours) and existing related works.} In terms of the given language prompt, NuGrounding can detect the described objects even if they contain various challenges, like crossing different views and occluded.}
    \label{fig:vis}
\end{figure*}

\textbf{Implementation Details.} 
We choose llava-1.5-7B \cite{llava-1.5} as our base multi-modal LLM with its official pre-training weight. The dimension of LLM last hidden layer $C_L$ is 4096. During training stage, we use LoRA \cite{lora} to finetune the LLM to preserve its original instruction understanding ability. StreamPETR \cite{streampetr} is chosen as our BEV-based detector and use its official 3D object detection pretraining. The image backbone are V2-99 \cite{vovnet} and ViT \cite{ViT}. The number and dimension of object query is 900 and 256, respectively. We use AdamW \cite{AdamW} optimizer with learning rate, batch size, and epoch set to 1e-5, 1, 1.

\subsection{Comparison with State-of-the-art Methods}
Since there are no existing methods for the multi-view 3D visual grounding task and dataset, we adapt the MLLM-based 3D scene understanding work, i.e., ELM \cite{elm}, NuPrompt \cite{nuprompt} and OmniDrive \cite{omnidrive} on our proposed NuGrounding dataset. As shown in Tab. \ref{tab:compareSOTA}, our proposed method using a small image backbone, V2-99 \cite{vovnet} outperforms all the previous SOTA methods. Furthermore, for efficiency, we only adopt ViT-B rather than the larger ViT-L as backbone, which yet achieves 0.59 precision, 0.64 recall, 0.40 mAP and 0.48 NDS for the average of four levels, which leading NuPrompt 0.30, 0.35, 0.29 and 0.26, respectively. 
% Besides, results show that ViT backbone brings a notable improvement than V2-99. It is because that ViT adopts the transformer architecture as MLLM, which lead to a pure transformer framework \cite{vilt}. 
In summary, benefiting from our 3D visual grounding framework and meticulously designed modules, our method outperforms other counterparts.

\subsection{Ablation Study}
\label{sec:Ablation Study}
In ablation studies, all experiments are conducted on the subset (attributes include category and relationship) of our NuGrounding dataset, and the image backbone is V2-99 to save computational resource if not specifically mentioned.

\textbf{Paradigm Design}
% Our attractive characteristic is the multi-view 3D visual grounding framework, which combines the instruction understanding capabilities of MLLM with the precise localization abilities of specialist detection model. 
To assess the impact of the components of the framework, we use textual thinking, including number, category, and relationship as our baseline method. As shown in Tab. \ref{tab:components}, it laid a solid foundation and achieves 0.387 mAP and 0.445 NDS. Along with the context query, the mAP and the NDS metrics increase 0.056 and 0.042, which indicates that the decoupled task token and context token could enable more effective aggregation of multi-modal information. Additionally, our query selector results in some progress and the specific chosen number could be found in the independent ablation study of Tab. \ref{tab:QuerySelector}.

\textbf{Scene Representation.}
As a comparison to our object query input, we employ blind, which means empty visual embeddings, and bev feature as the visual input. Results in Tab. \ref{tab:scene_representation} demonstrate a clear advantage of the object query input. This is because that object query can capture 3D geometric priors information from 3D detector. Obviously, the blind result reveals the interaction between the visual embeddings and the LLM tokens is necessary.

\textbf{Textual Thinking.}
We perform an ablation study to assess the impact of textual thinking. As presented in Tab. \ref{tab:context_learning}, with the growth of token length and detail information, the effect improves continuously. These results indicates that the detailed textual response enables the MLLM to take more time and more response tokens for thinking and increase the model complexity and representation ability.

\textbf{Query Selector.}
We vary the number of object queries from 32 to 900 (full), to investigate its impact on performance. As shown in Tab. \ref{tab:QuerySelector}, 256 queries are better than 900 queries, which indicates that we select the most semantically relevant object queries and mitigate irrelevant object noise. However, the results of 32 and 64 queries show that a reasonable number of queries is essential.

\subsection{Qualitative Results.}
As depicted in Fig. \ref{fig:vis}, we provide a visual comparison with existing related works, including NuPrompt \cite{nuprompt} and OmniDrive \cite{omnidrive}. These models struggle to handle the displayed cars with both false positive and false negative.  Especially, the Nuprompt misses the silver car behind the red car and the OmniDrive fails to catch the black car in front of the red car. It is worth noting that our approach not only generates accurate textual responses, but also produces precise and high-quality 3D grounding results.

\section{Conclusion}
In this work, we have presented NuGrounding, the first large-scale benchmark for multi-view 3D visual grounding in autonomous driving. We have designed the Hierarchy of Grounding (HoG) method to construct a diversity, scalability, and generalization dataset. To deal with the multi-view 3D visual grounding task of the dataset, we have proposed a novel framework that leverages the instruction comprehension abilities of multi-modal LLMs and the precise localization abilities. We have verified the effectiveness of our models and the adapted baselines, demonstrating their remarkable generality and flexibility.

{
    \small
    \bibliographystyle{ieeenat_fullname}
    \bibliography{main}
}
\newpage
\clearpage

\appendix

\newcounter{counter}[section]
\twocolumn[{%
\renewcommand\twocolumn[1][]{#1}
\begin{center}
    \Large
    \textbf{Supplementary Material}
    \\[20pt]
\end{center}
}]

\section{A Sample of HoG Motivation}
\label{sec:HoG_motivation}

\begin{figure*}[t]
    \centering
    \includegraphics[width=1.0\linewidth]{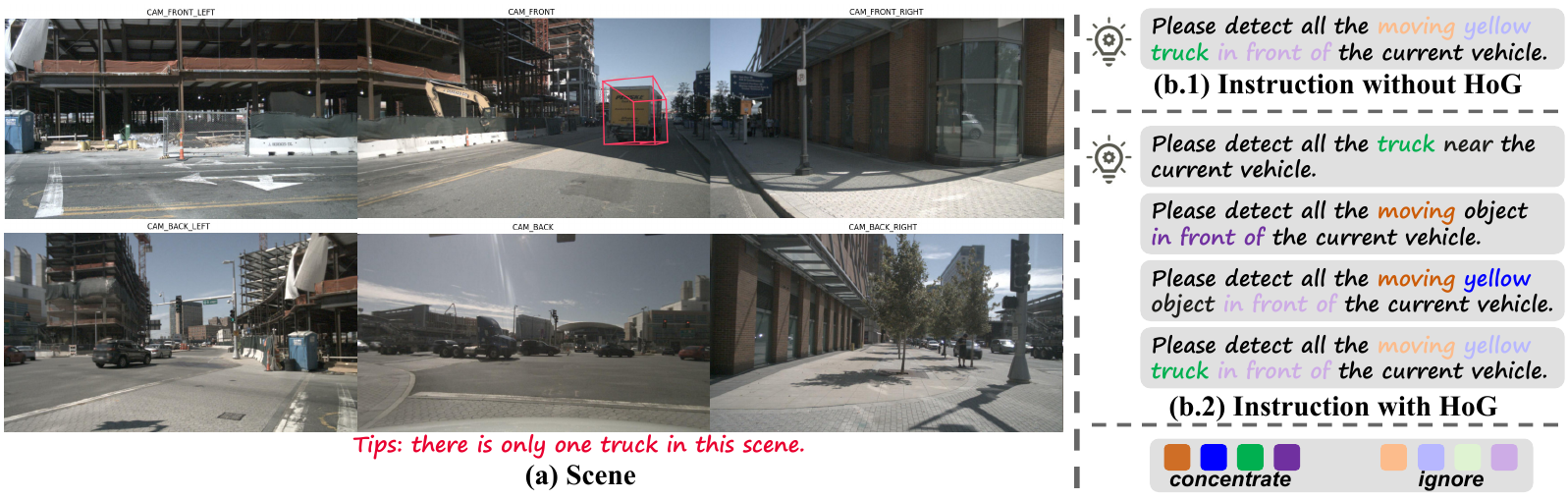}
    \caption{
    \textbf{A sample of HoG motivation.} \textbf{(a)} a driving scene where there is only one truck. \textbf{(b.1)} As there is only one truck, models can locate the target object by concentrating solely on the object category information \textit{``truck"} in the instruction without HoG. Training on such samples would guide the model in taking shortcuts to ignore other information. \textbf{(b.2)} Instruction with HoG helps models to concentrate on various useful attributes, which enhances the generalization capability for complex human instruction patterns.}
    \label{fig:HoG_sample}
\end{figure*}

As mentioned in Sec. \ref{sec:HoG} in the main paper, expressions coupling all four attribute types without hierarchy may lead to inductive bias, which motivates us to present the Hierarchy of Grounding (HoG) method to construct our NuGrounding dataset. Here we provide a sample visualization to explain it, as shown in Fig. \ref{fig:HoG_sample}.

\section{Loss Function}
Our method is trained under one single stage, where each sub-task could gradually learn to find out its own optimization direction. There are LLM text generation loss $\mathcal{L}_{txt}$, detection loss $\mathcal{L}_{det}$ and contrastive loss for query selection $\mathcal{L}_{c}$. And we adjust weights to make sure all terms have the same magnitude, as in the following equation:
\begin{gather}
\mathcal{L}={w}_{txt}\mathcal{L}_{txt} + {w}_{det}\mathcal{L}_{det} + {w}_{c}\mathcal{L}_{c}
\end{gather}

Specifically, the $\mathcal{L}_{txt}$ is the auto-regressive cross-entropy loss for LLM text generation, and the corresponding tensor of \texttt{[EMB]} token is excluded from calculating this loss. The $\mathcal{L}_{det}$ includes focal loss for classification and L1 loss for bounding box regression. As for the $\mathcal{L}_{c}$, we follow \cite{groundingdino} to compute focal loss for each logit in similarity matrix.

\section{Evaluation Metrics}
We follow most previous work on 3D object detection \cite{streampetr, bevdepth, bevdet, bevformer} to adopt mAP and NDS as metrics. mAP is the well-known mean Average Precision metric: for a given match threshold we calculate average precision (AP) by integrating the recall-precision curve and finally compute the mean across classes. 
NDS is a specific metric from NuScenes \cite{nuscenes}, which consolidates the metrics by computing a weighted sum of mAP and all kinds of object attribution errors (average translation error, average scale error, average orientation error, average velocity error, average attribute error). Note that it is difficult to depict the Precision-Recall curve of the text-based methods without predicting the confidence score, so we report exact Precision and Recall metrics for fair comparison. To calculate the precision and recall of our methods, we set the confidence score and central distance thresholds to 0.25 and 2$m$.

\section{Additional Experiments}

\subsection{Oracle Comparisons}
\label{sec:oracle}

\begin{table}[ht]
    \centering
    \resizebox{0.35\textwidth}{!}{
    \begin{tabular}{c|cc}
    \toprule[1.25pt]
    Model & mAP & NDS \\
    \midrule
    Ours (Movement) & 0.369 & 0.495 \\
    Ours (Relationship) & 0.414 & 0.478 \\
    Ours (Category) & 0.446 & 0.497 \\
    Ours (Mean) & 0.410 & 0.490 \\
    \hdashline
    Ours (Object) & 0.351 & 0.415 \\
    \midrule
    Specialist & 0.472 & 0.563 \\ 
    \bottomrule[1.25pt]
    \end{tabular}               
    }
    \caption{\textbf{Oracle comparisons.} Ours (Movement, Relationship, Category) indicates that the object detection results are assembled from level-1 subset of NuGrounding, and we calculate the mean average of these three results as Ours (Mean). Ours (Object) indicates that we directly generate the detection results by asking the out-of-distribution instruction \textit{``Please detect all the object near the current vehicle"}. Specialist indicates the results of specialist object detection models, which can not comprehend instructions.}
    \label{tab:oracle}
\end{table}

Our framework combines the instruction comprehension abilities of multi-modal LLMs with the object localization abilities of specialist detection models. It is valuable to evaluate the localization precision gap between our combined framework and oracle detection models.

However, the ground truth objects of NuGrounding dataset cannot perfectly cover the target objects of the detection task. Therefore, Since the first-level subsets of NuGrounding can be assembled together to form the detection task, we utilize these subsets for evaluation. For example, we assemble the prediction objects of instruction \textit{``Please detect all the moving object near the current vehicle"} together with the prediction objects of instruction \textit{``Please detect all the stopped object near the current vehicle"} to form the detection results, indicated as Ours (Movement) in Tab. \ref{tab:oracle}. The tiny gap between Ours (Mean) and specialist proves the effectiveness of our seamless combination of MLLMs and specialist models.

In addition, we generate the detection results by directly asking the out-of-distribution instruction \textit{``Please detect all the object near the current vehicle"} which does not belong to the training set, indicated as Ours (Object) in Tab. \ref{tab:oracle}. The result shows that our model can generalize to unseen instruction patterns to locate the instruction-following objects.

\subsection{Data Scaling Analysis}
\label{sec:data_scaling}

\begin{figure}[ht]
    \centering
    \includegraphics[width=0.45\textwidth]{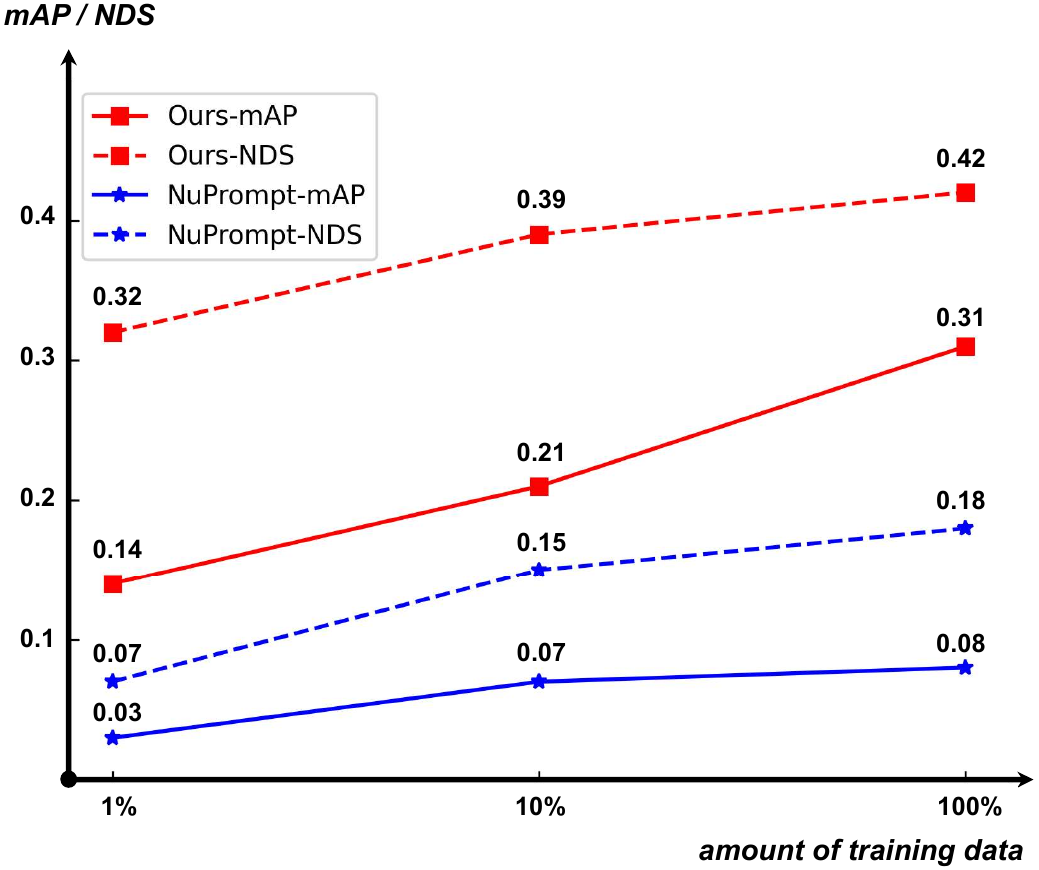}
    \caption{
    \textbf{Data scaling analysis.} The horizontal axis represents the amount of training data, which is uniformly sampled from the entire NuGrounding dataset.
    }
    \label{fig:data_scaling}
\end{figure}

Understanding how performance scales with varying data sizes is important for practical applications, especially in data-scarce environments. As shown in Fig. \ref{fig:data_scaling}, We uniformly sample 1\% and 10\% of the entire NuGrounding dataset for training, and report the average mAP and NDS results of our framework and NuPrompt \cite{nuprompt}. The results show that our framework efficiently learns from limited training data, and ours on 1\% training data even outperforms NuPrompt on 100\% training data. Besides, the mAP of NuPrompt on 10\% data and on 100\% data is similar, indicating that NuPrompt has approached saturation even with limited data. In contrast, the growth curve of ours-mAP suggests the big data scaling ability of our model.

\section{Visualizations}

\subsection{Qualitative Comparisons}

\begin{figure*}[t]
    \centering
    \includegraphics[width=1.0\linewidth]{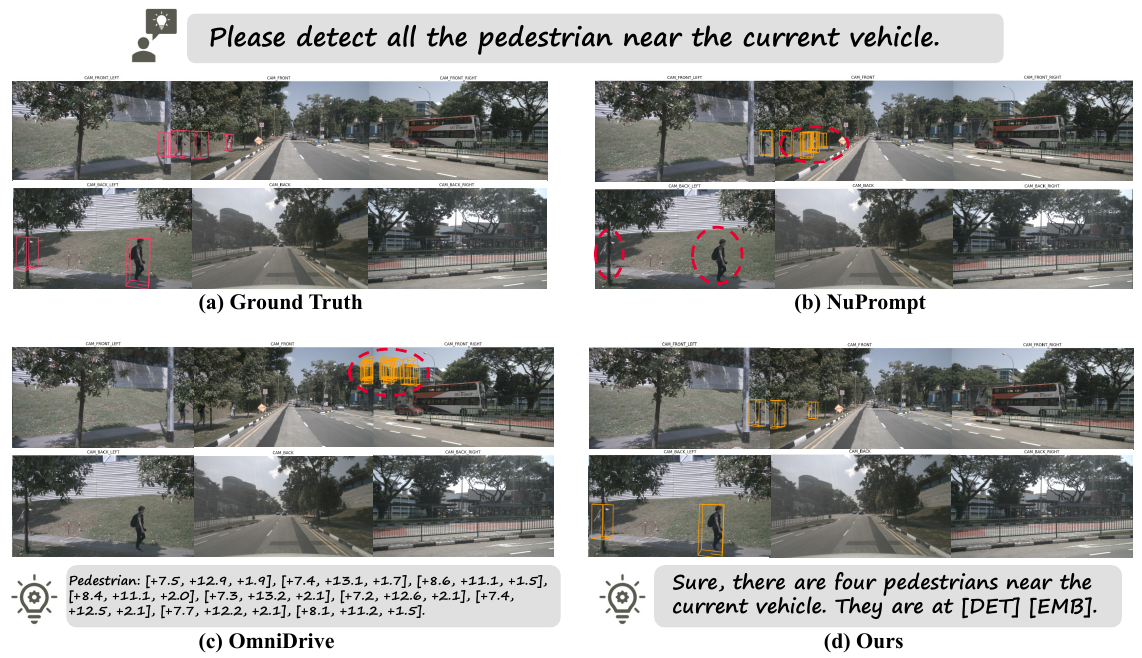}
    % \vspace{-2.0em}
    \caption{
    \textbf{Visual comparison on NuGrounding level-1 subset.}}
    \label{fig:level-1}
\end{figure*}

\begin{figure*}[t]
    \centering
    \includegraphics[width=1.0\linewidth]{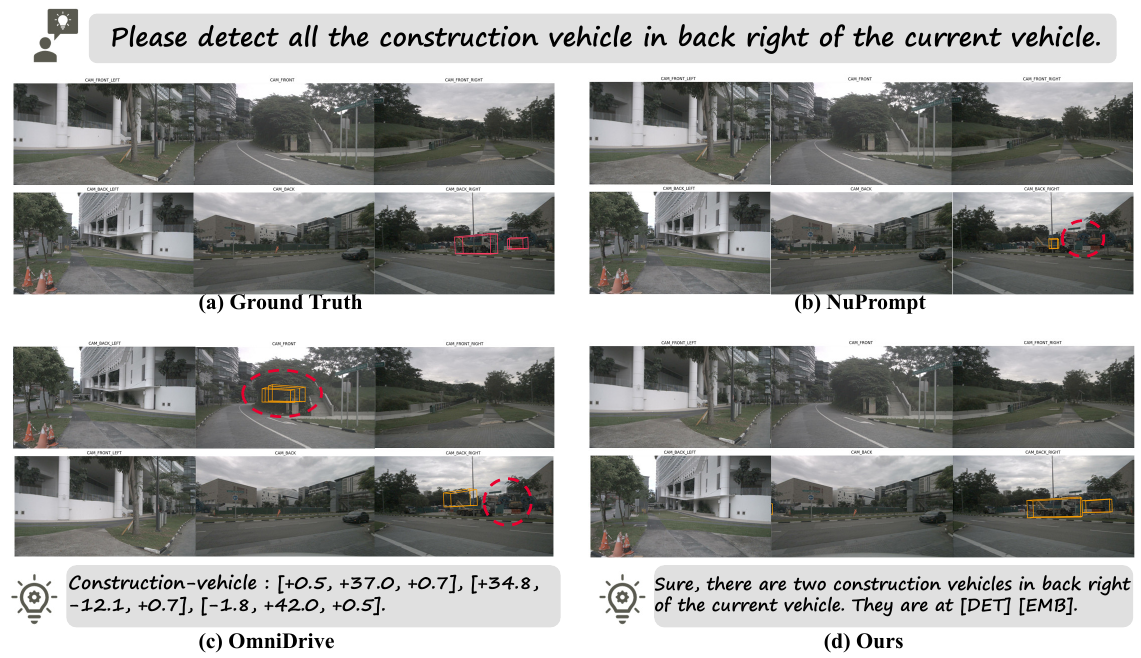}
    % \vspace{-2.0em}
    \caption{
    \textbf{Visual comparison on NuGrounding level-2 subset.}}
    \label{fig:level-2}
\end{figure*}

\begin{figure*}[t]
    \centering
    \includegraphics[width=1.0\linewidth]{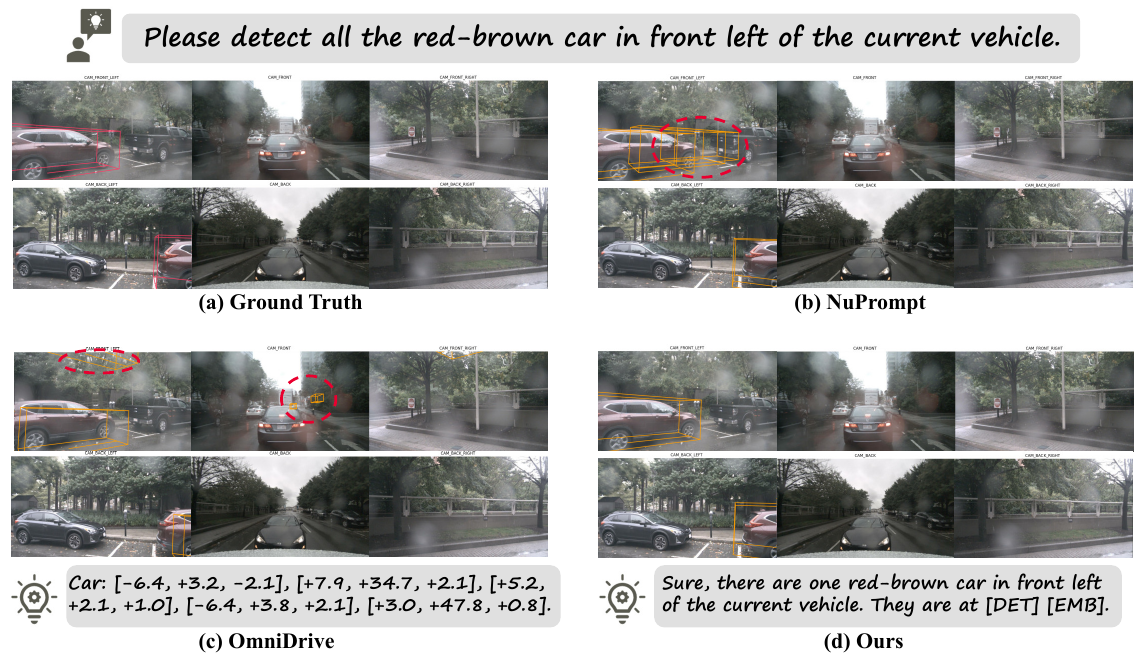}
    % \vspace{-2.0em}
    \caption{
    \textbf{Visual comparison on NuGrounding level-3 subset.}}
    \label{fig:level-3}
\end{figure*}

\begin{figure*}[t]
    \centering
    \includegraphics[width=1.0\linewidth]{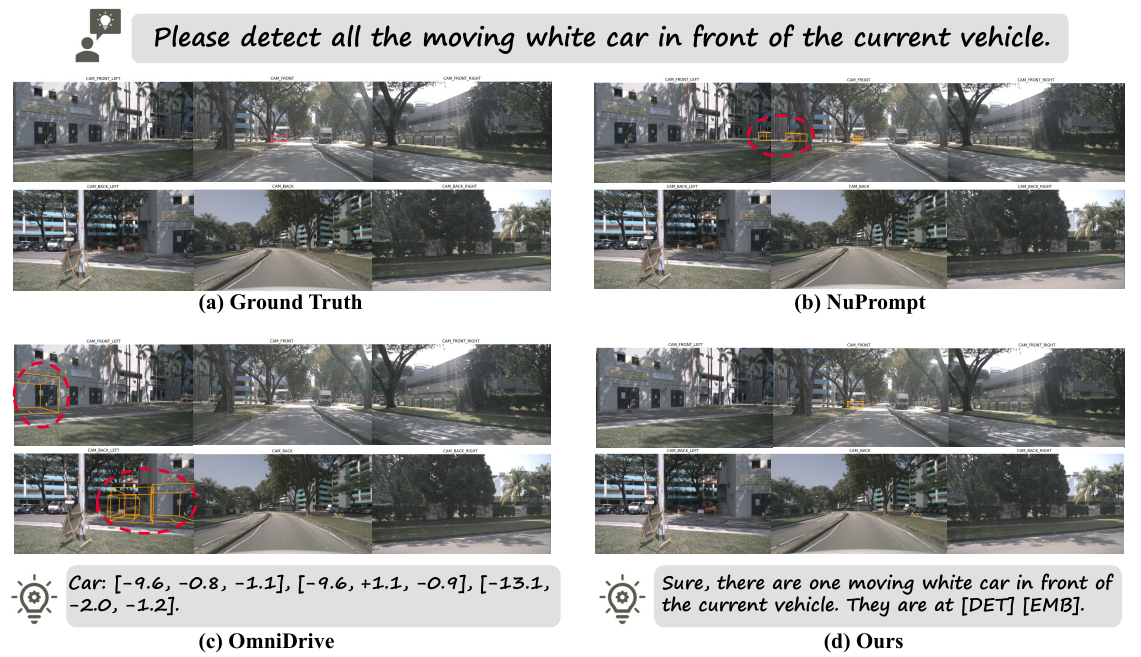}
    % \vspace{-2.0em}
    \caption{
    \textbf{Visual comparison on NuGrounding level-4 subset.}}
    \label{fig:level-4}
\end{figure*}

As shown in Fig. \ref{fig:level-1}-\ref{fig:level-4}, we provide more visual comparisons among our framework and existing related works on each level of NuGrounding dataset. It can be observed that our framework faithfully adheres to human instructions and predicts precise bounding boxes for each level, while other methods struggle to concentrate on target objects and locate objects precisely.

\subsection{Failure Case}

\begin{figure*}[t]
    \centering
    \includegraphics[width=1.0\linewidth]{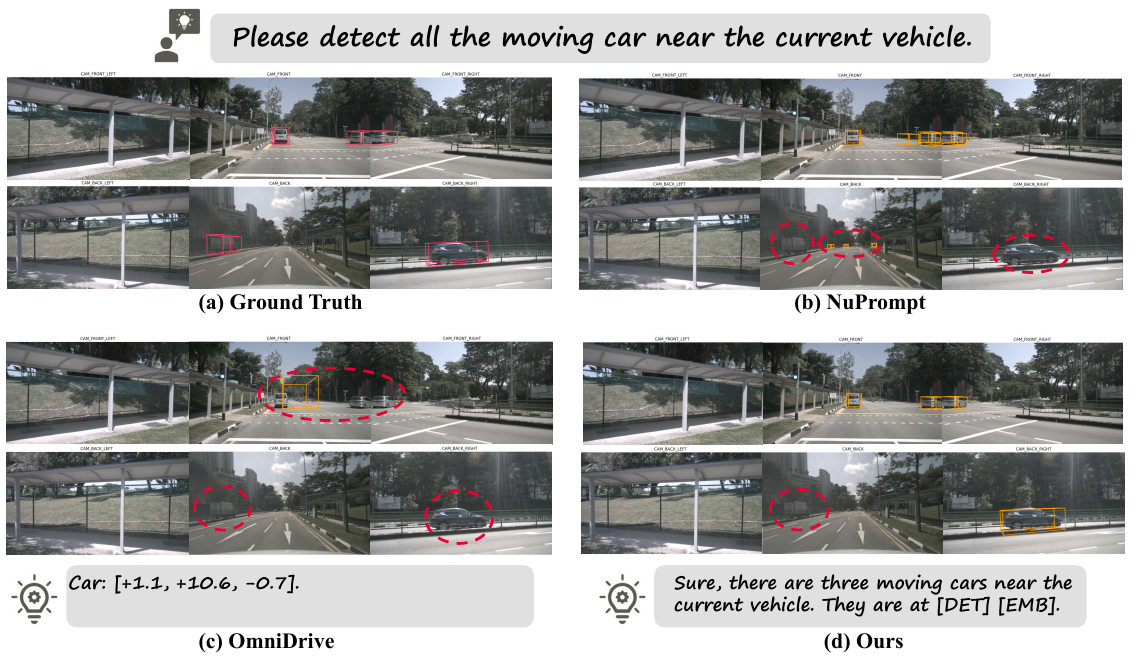}
    % \vspace{-2.0em}
    \caption{
    \textbf{Failure case.} Ours misses a moving car in the CAM BACK.}
    \label{fig:failure_case}
\end{figure*}

We show the grounding results of a challenging scene in Fig. \ref{fig:failure_case}. Our framework misses a moving car, while other related methods also fail to detect it. We think that the hard threshold 0.3$m/s$ of movement attribute collections brings about some wrong annotations. Some objects are actually moving slowly but are annotated as stopped. These cases will mislead the model learning procedure when facing slowly moving objects, resulting in false negative detections.

\section{Future Outlook}
Our framework which combines the MLLMs with the query-based methods could be extended to other tasks (e.g. open vocabulary 3D visual grounding, instruction-guided planning and so on). We are looking forward to do these in the future.

\end{document}